\documentclass[11pt]{article}

\usepackage[final]{acl}

\usepackage{times}
\usepackage{latexsym}
\usepackage{amsmath}
\usepackage{cleveref}
\usepackage{longtable}
\usepackage{tcolorbox}
\usepackage{adjustbox}
\usepackage[T1]{fontenc}

\usepackage[utf8]{inputenc}
\usepackage{makecell}

\usepackage{xcolor}
\definecolor{english}{HTML}{FFC0CB}
\definecolor{revf}{HTML}{F27E9D}
\definecolor{revp}{HTML}{592A37}
\definecolor{shuf3}{HTML}{77D9CF}
\definecolor{shuf5}{HTML}{5AB8C6}
\definecolor{shufeo}{HTML}{85AFEB}
\definecolor{shuf10}{HTML}{05A6A6}
\definecolor{shufd}{HTML}{144045}
\definecolor{shufnd}{HTML}{666666}
\usepackage{graphbox} %
\usepackage{microtype}
\usepackage{xurl}

\usepackage{inconsolata}

\usepackage{graphicx}
\usepackage{subfig}
\usepackage[table]{xcolor}
\usepackage{booktabs}
\usepackage{array}

\usepackage{titlesec}

\titlespacing*{\paragraph}{0pt}{0.6ex plus 0.3ex minus 0.2ex}{0.5em}

\definecolor{hlPink}{HTML}{FFD1DC} 
\definecolor{hlBlue}{HTML}{E0F7FA}
\definecolor{hlGreen}{HTML}{E8F5E9} 
\definecolor{hlYellow}{HTML}{FFF9C4} 
\definecolor{hlPurple}{HTML}{F3E5F5} 
\definecolor{hlOrange}{HTML}{FFE0B2} 
\definecolor{hlCyan}{HTML}{FFCCBC} 
\definecolor{hlGray}{HTML}{F5F5F5} 

\newcommand{\hl}[2]{%
  \setlength{\fboxsep}{1pt}%
  \colorbox{#1}{\strut #2}%
  \hspace{1.5pt}%
}

\newcommand{\sqR}{%
  \begingroup
  \setlength{\fboxsep}{1.2pt}%
  \setlength{\fboxrule}{0.4pt}%
  \fbox{\scriptsize\sffamily R}%
  \endgroup
}

\title{When transformers learn ``impossible'' languages, what do they learn?}

\author{Ram Janarthan$^*$~~~Coleman Haley$^{*, \dagger}$~~~Sharon Goldwater\\
University of Edinburgh \\
\texttt{$^\dagger$coleman.c.haley@gmail.com}
}

\usepackage{todonotes}
\begin{document}
\maketitle
\begin{abstract}

Recent work suggests that transformer language models show a bias towards human languages over unnatural (``impossible'') languages argued to be unacquirable by humans. However, this literature has largely based these claims on differences in sample efficiency and test-set perplexity, rather than on direct evaluations of the linguistic capacities that could plausibly explain non-attestation in human languages. We evaluate two theoretically motivated linking hypotheses: impossibility arising from deficiencies in {\em grammatical sensitivity} or {\em generative production}. Using GPT-2 style models trained on perturbed ``impossible'' variants of English, we measure sensitivity to grammaticality using BLiMP minimal pairs, finding that model performance exhibits only gradual degradation, mediated by the language's information locality. In contrast, these models exhibited pronounced failures in generation, producing substantially fewer high-quality sentences at longer lengths. Together, these results suggest generative deficiency and transmission failures as a plausible linking hypothesis between language model behaviour and non-attestation of impossible languages.
\end{abstract}

\section{Introduction}
\def\thefootnote{*}\footnotetext{These authors contributed equally to this work.}
\def\thefootnote{\arabic{footnote}}

\setcounter{footnote}{0}

Recent successes of language models (LMs) have sparked a debate over whether they can inform theories of human language acquisition. LMs appear to exhibit substantial linguistic competence, producing novel grammatical sentences \citep{mccoy-et-al-2023-how} and showing sensitivity to diverse linguistic phenomena \citep{hu-et-al-2023-prompting, linzen-et-al-2021-syntactic, wilcox-et-al-2024-using}---leading some to suggest that general-purpose learning mechanisms may suffice to acquire language \citep{mahowald-et-al-2024-dissociatinga, futrell-et-al-2025-how}. An opposing perspective instead posits that language acquisition is guided by innate, language-specific constraints that sharply restrict the space of possible human languages, accounting for both cross-linguistic universals and the systematic absence of many logically possible but unattested languages \citep{chomsky-1966-explanatory, chomsky-1998-nature, moro-2008-boundaries}. Under this view, LMs, lacking language-specific biases, have been claimed to be irrelevant for explaining human language acquisition and patterns of unattested languages \citep{chomsky-et-al-2023-opinion, moro-et-al-2023-large}.

Responding to these claims, \citet{kallini_mission_2024} performed an empirical study showing that transformer LMs trained on a developmentally-plausible amount of data learned English better and more quickly than modified English variants with properties argued to be impossible (henceforth ``impossible languages''). Subsequently, these authors, in \citet{kallini-et-al-2025-languagea} and \citet{futrell-et-al-2025-how}, have argued that the domain-general inductive biases present in LMs can inform linguistic theory by modelling the pattern of \mbox{(un)at}tested human languages. 
These arguments have relied on differences in learning dynamics: LMs trained on impossible languages tend to converge more slowly and yield higher test-set perplexity than those trained on natural languages. This leaves the {\em linking hypothesis} to the non-attestation of these languages under-developed\footnote{Indeed, \citet{kallini-et-al-2025-languagea} have subsequently called for the development of stronger linking hypotheses between this research program in language models and human language.}:
 why should differences in perplexity or sample efficiency correspond to {\em exclusion} from the space of extant human languages?

In this study, we focus on two kinds of linguistic capacities as candidate linking hypotheses between LM behaviour and linguistic “impossibility,” motivated by contrasting theoretical orientations in linguistics. \textbf{Sensitivity to grammatical well-formedness}  plays a central role in accounts that attribute non-attestation to limits on what learners can infer from input \citep{chomsky-1980-rules, chomsky-1966-explanatory}, while reliable \textbf{generation of well-formed sentences} is emphasized by accounts that locate impossibility in failures of transmission across speakers \citep{kirby-et-al-2008-cumulative}. If models fail to acquire sensitivity to some critical aspects of a language, then impossibility might arise from unlearnability; while if models cannot generate well-formed sentences in the language, impossibility might reflect difficulties in transmitting the language. %

We focus our study on the capacities acquired by transformer models trained on a cognitively plausible amount of data: either the English BabyLM corpus \citep{warstadt-et-al-2023-findings} or impossible variants defined  by \citet{kallini_mission_2024} as permutations of the English sentences. 
To evaluate {\em grammatical sensitivity}, we use the BLiMP minimal pair dataset \citep{warstadt-et-al-2020-blimp}, forming ``impossible'' variants of BLiMP by permuting the BLiMP stimuli in the same manner as the BabyLM data was permuted by \citet{kallini_mission_2024}. To evaluate {\em generative performance}, we leverage the fact that grammatical strings of \citeauthor{kallini_mission_2024}'s impossible languages can be inverted to produce grammatical strings of English. Generating sentences from each impossible language model and converting them to their English equivalents, we use an LLM to evaluate the quality of the generations.

We find models trained on 
impossible languages acquire substantial {\em passive grammatical sensitivity}, degrading gradually in performance with respect to both overall test-set perplexity and $m$-local entropy, a measure of information locality proposed to explain asymmetries in impossible language learning \cite{someya-et-al-2025-information}. This result stands in contrast to claims that human learners would fail to acquire grammatical competence from such data.\footnote{For example, \citet{chomsky-1980-rules} outlines how human learners would be \emph{unable} to acquire a rule based on linear ordering, as opposed to hierarchical structure.} %

In terms of {\em generative performance}, we find that models trained on impossible languages produce substantially fewer well-formed sentences in their languages, in a manner which does {\em not} strictly align with held-out test set perplexity or $m$-local entropy.

Together, these results suggest that language models acquire substantial (if slightly degraded) {\em grammatical sensitivity} to impossible languages, but tend to fail at {\em generative performance}, suggesting generative performance as a potential linking hypothesis between poorer LM distributional modelling of impossible languages and human non-occurrence.
While these results will not resolve the debate about whether transformers are good models of human learners, they do provide a potential explanation for {\em why} some languages could be impossible, even for a learner with no strong language-specific inductive biases. %
Our study also provides a more principled methodology for studying impossibility with language models, which we encourage future studies to adopt and extend.\footnote{The code for this paper is available at: \url{https://github.com/ramjanarthan/impossible-languages}} 

\begin{table*}[t]
\begin{adjustbox}{width=\textwidth,center}

\fontsize{10}{10.5}\selectfont
    \centering
    \begin{tabular}{ll>{\raggedright}p{5.1cm}l}
    \toprule
    \textbf{Language} &  \textbf{Abbr.} & \textbf{Perturbation Rule} & \textbf{Example Sentence} \\
    \midrule
    \tcbox[colback=english,colframe=english,top=0pt,left=0pt,right=0pt,bottom=0pt,box align=base]{\textsc{English}} & E & No perturbation &
    \hl{hlPink}{Jessica}\hl{hlBlue}{ stole}\hl{hlGreen}{ this}\hl{hlYellow}{ rabbit}\hl{hlPurple}{'s}\hl{hlOrange}{ hat}\hl{hlCyan}{.} \vspace{4pt}  \\

    \tcbox[colback=revf,top=0pt,left=0pt,right=0pt,bottom=0pt, colframe=revf,box align=base]{\textsc{FullReverse}} & FR & Randomly insert a special \sqR\space token, and reverse the ordering of all tokens  &
    \hl{hlCyan}{.}\hl{hlOrange}{ hat}\hl{hlPurple}{'s}\hl{hlYellow}{ rabbit}\hl{hlGreen}{ this}\hl{hlGray}{\sqR}\hl{hlBlue}{ stole}\hl{hlPink}{Jessica}  \\

\tcbox[colback=revp,top=0pt,left=0pt,right=0pt,bottom=0pt,colframe=revp,box align=base]{\textcolor{white}{\textsc{PartialReverse}}} & PR & Randomly insert a special \sqR\space token, and only reverse the ordering of all tokens following it &
    \hl{hlPink}{Jessica}\hl{hlBlue}{ stole}\hl{hlGray}{\sqR}\hl{hlCyan}{.}\hl{hlOrange}{ hat}\hl{hlPurple}{'s}\hl{hlYellow}{ rabbit}\hl{hlGreen}{ this}  \\

    \tcbox[colback=shuf3,colframe=shuf3,top=0pt,left=0pt,right=0pt,bottom=0pt,box align=base]{\textsc{LocalShuffle3}} & S3 & Deterministically shuffle tokens within a window of size $3$ &
    \hl{hlGreen}{ this}\hl{hlPink}{Jessica}\hl{hlBlue}{ stole}\hl{hlOrange}{ hat}\hl{hlYellow}{ rabbit}\hl{hlPurple}{'s}\hl{hlCyan}{.}  \\

    \tcbox[colback=shuf5,colframe=shuf5,top=0pt,left=0pt,right=0pt,bottom=0pt,box align=base]{\textsc{LocalShuffle5}} & S5 & Deterministically shuffle tokens within a window of size $5$ &
    \hl{hlGreen}{ this}\hl{hlPurple}{'s}\hl{hlYellow}{ rabbit}\hl{hlPink}{Jessica}\hl{hlBlue}{ stole}\hl{hlOrange}{ hat}\hl{hlCyan}{.}  \\
    \tcbox[colback=shufeo,colframe=shufeo,top=0pt,left=0pt,right=0pt,bottom=0pt,box align=base]{{\textsc{EvenOddShuffle}}} & SEO & Reorder tokens so even-indexed tokens appear before odd-indexed tokens & 
    \hl{hlPink}{Jessica}\hl{hlGreen}{ this}\hl{hlPurple}{'s}\hl{hlCyan}{.}\hl{hlBlue}{ stole}\hl{hlYellow}{ rabbit}\hl{hlOrange}{ hat}  \\

    \tcbox[colback=shuf10,colframe=shuf10,top=0pt,left=0pt,right=0pt,bottom=0pt,box align=base]{\textsc{LocalShuffle10}} & S10 & Deterministically shuffle tokens within a window of size $10$ &
    \hl{hlGreen}{ this}\hl{hlPurple}{'s}\hl{hlYellow}{ rabbit}\hl{hlCyan}{.}\hl{hlOrange}{ hat}\hl{hlPink}{Jessica}\hl{hlBlue}{ stole}  \\
    \tcbox[colback=shufd,colframe=shufd,top=0pt,left=0pt,right=0pt,bottom=0pt,box align=base]{\textcolor{white}{\textsc{DetermShuffle}}} & DS &Deterministically shuffle all tokens, with shuffling seed $21$ &
    \hl{hlPink}{Jessica}\hl{hlPurple}{'s}\hl{hlBlue}{ stole}\hl{hlOrange}{ hat}\hl{hlYellow}{ rabbit}\hl{hlCyan}{.}\hl{hlGreen}{ this}  \\
    \tcbox[colback=shufnd,colframe=shufnd,top=0pt,left=0pt,right=0pt,bottom=0pt,box align=base]{\textcolor{white}{\textsc{NondetermShuffle}}} & NDS & Nondeterministically shuffle all tokens &
    \hl{hlGreen}{ this}\hl{hlPurple}{'s}\hl{hlYellow}{ rabbit}\hl{hlCyan}{.}\hl{hlOrange}{ hat}\hl{hlPink}{Jessica}\hl{hlBlue}{ stole}  \\
    \bottomrule
    \end{tabular}
\end{adjustbox}
    \caption{Examples of sentences in English and impossible languages. Coloured boxes represent GPT-2 tokens. We use the same colors and abbreviations to represent languages throughout the paper. Languages in the table are ordered by increasing 4-local entropy (see Section~\ref{sec:background}).}
    \label{tab:examples_imp_lang}
\end{table*}

\section{Background}
\label{sec:background}

What kinds of languages  are ``impossible'' is the subject of ongoing discussion among linguists, due to both the difficulty in establishing true universals of natural languages, and uncertainty about whether unattested languages could in principle be learned. Nevertheless, \citet{kallini_mission_2024} proposed a set of candidate impossible languages by defining perturbations of the word order of a sentence applied to a natural language (English). These perturbations manipulate sentences in ways that are never observed in any human languages and have been previously hypothesized to be impossible \citep{moro-2008-boundaries, mitchell-et-al-2020-priorless} due to their use of unattested and ``unnatural'' word orders. Each perturbation is applied to a tokenized sentence as described in Table \ref{tab:examples_imp_lang}, and all except one can be deterministically converted to English equivalents, a fact which we exploit in Section~\ref{sec:generation}.
This deterministic mapping also means that the true entropy of these languages is the same as the entropy of English, so a model that successfully learns both English and the impossible languages should have the same perplexity on each.

In fact, this is not what \citet{kallini_mission_2024} found. They trained GPT-2 transformer models \citep{radford2019language} on both the BabyLM corpus \cite{warstadt-et-al-2023-findings} and perturbed (impossible) versions of it, and showed that the models trained on impossible languages converged more slowly and ended up with slightly higher perplexity on held out data, suggesting more difficulty learning the impossible languages. However, results based on perplexity alone are somewhat difficult to interpret (how much worse constitutes a learning failure?), and follow-ups on other languages have shown more mixed results \cite{ziv-etal-2026-biasless, yang-et-al-2025-anything}. More importantly, there is no theoretically-motivated link between  low perplexity and impossibility.  In Section~\ref{sec:linking}, we explore two more theoretically-elaborated hypotheses linking LM performance to impossibility.\footnote{We note that \citet{xu-et-al-2026-can} also evaluated models' grammatical sensitivity using minimal pairs, but they did so using typologically {\em implausible} (rather than impossible) languages, making their results less suitable for  determining a linking hypothesis between impossibility and LM performance.}

While the primary aim of this study is to test these more explicit linking hypotheses, we also secondarily study how our results relate to two prior measures of impossibility: perplexity and $m$-local entropy. \citet{someya-et-al-2025-information} proposed $m$-local entropy as an information-theoretic measure characterizing impossible languages. They defined $m$-local entropy as the next-symbol entropy given a context of size $m$$-1$, and estimated it for these languages using \textit{n}-gram models trained on perturbed corpora very similar to those in \citet{kallini_mission_2024}. 
They showed a strong positive correlation between $m$-local entropy (strongest when $m=4$) and the perplexity of transformer models trained on different impossible languages, concluding that transformers exhibit an information-locality bias that drives \citeauthor{kallini_mission_2024}'s hierarchy of impossibility.

\begin{table*}[h]
\renewcommand{\arraystretch}{1.3}
    \centering
        \begin{tabular}{lcc}
         \toprule
    \textbf{Language} & \textbf{Grammatical Example} & \textbf{Ungrammatical Example} \\
    \midrule
             \textsc{English} & Rodney goes to \textbf{this} new mall. & Rodney goes to \textbf{these} new mall. \\
             \textsc{FullReverse} & . mall new \textbf{this} to goes\sqR neyRod & . mall new \textbf{these} to goes\sqR neyRod \\
             \textsc{LocalShuffle3} &  goesRodney new to \textbf{this} mall. &  goesRodney new to \textbf{these} mall. \\
             \textsc{NondetermShuffle} &  goesRod new. toney \textbf{this} mall &  goesRod new. toney \textbf{these} mall \\
            \hline
        \end{tabular}%
    \caption{Examples of a minimal pair in different (impossible) languages for the BLiMP task ``Determiner Noun Agreement with Adjective''.}
    \label{tab:impossible_blimp_examples}  
\end{table*}

\section{Linking Hypotheses}\label{sec:linking}

Prior work on impossible languages on LMs has focused on learning dynamics and held-out perplexity, implicitly treating poorer compression as an explanation for, or a diagnostic of, impossibility. In this section, we argue that this move is theoretically underjustified. We propose two alternative linking hypotheses arising from opposing theoretical orientations within linguistics: deficiencies in {\em grammatical sensitivity} and {\em generative performance}, describing how they may diverge from perplexity and each other.

\subsection{Grammatical sensitivity in LMs}\label{sec:temperature}
Generative linguists have generally put substantial emphasis on the challenge of acquiring language from the limited data humans are exposed to during development \citep{chomsky-1966-explanatory, chomsky-1980-rules}. For example, \citep{chomsky-1980-rules} claimed that the linguistic input available to learners is insufficient to determine the correct grammatical generalizations without strong, language-specific inductive biases. On this view, impossible languages are impossible because learners cannot reliably acquire the abstract grammatical distinctions required to discriminate grammatical from ungrammatical strings, even with extensive exposure \citep{moro-2008-boundaries}. If this claim is correct, then models trained on impossible languages should show enduring deficits in passive sensitivity to linguistic structure relative to models trained on natural languages, thereby explaining why these languages are unattested.

The most widely used and successful approach to evaluating sensitivity to grammatical well-formedness in LMs is the use of minimal pair datasets like BLiMP \citep{warstadt-et-al-2020-blimp}. Specifically, minimally-differing strings, one grammatical and one ungrammatical, are compared using perplexity, with the lower perplexity taken to be the model's preferred variant. The use of minimal pairs accounts for the ``noisy channel'' nature of sentence probability \citep{hu-et-al-2026-what,levy-2008-expectation}, which assigns lower values  to infrequent grammatical strings than to ungrammatical strings which are close to frequent grammatical strings.

\begin{figure*}
    \centering
    \hspace*{0.06\linewidth}\includegraphics[width=0.93\linewidth]{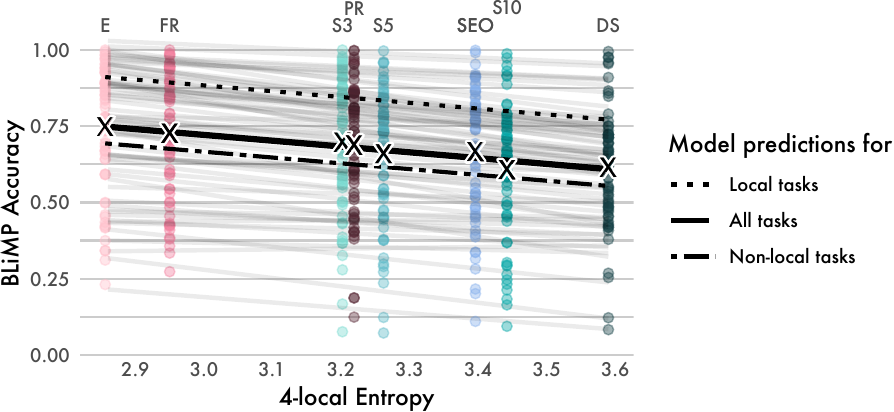}

    \caption{
    BLiMP task accuracy of models trained on English and impossible languages, correlated with 4-local entropy (see Appendix~\ref{app:perpplot} for the analogous plot of model accuracy against perplexity, which fits less well). Coloured dots and grey lines show accuracy and lines of best fit for individual tasks, with Xs indicating the mean for each language (labeled at top with abbreviations from Table~\ref{tab:examples_imp_lang}). The model indicates a modest, linear decline with increasing 4-local entropy (fitted black line). This decline is insensitive to task locality: tasks that are solved well by a 5-gram model in English exhibit almost the same rate of degradation as those that are not (dotted vs. dashed line).
    \label{fig:blimp}}
\end{figure*}

This minimal-pair view of grammatical sensitivity in language models displays
patterns which may diverge from the overall model loss. For example, consider the following family of Boltzmann distributions parameterized by temperature $T$:
\begin{equation}
P_T(x) = \frac{\exp(\operatorname{logits}(x)/T)}{\sum_{x^\prime\in V}\exp(\operatorname{logits}(x^\prime)/T)}\,,
\end{equation}
where $\operatorname{logits}(x)$ are the logits of some neural LM for an input $x$. While a language model most standardly refers to the distribution where $T=1$, in practice, generations frorm a LM are often drawn from a number of distributions in this family, with $T$ controlling how conservative the generations are.
These distributions have different entropies, which implies different perplexities on a test set, but varying $T$ is a rank-preserving operation: for any $x$ and $y$, $P_T(x)\! >\! P_T(y)$ implies $P_{T^\prime}(x)\! >\! P_{T^\prime}(y)$. That is, {\em the whole family of distributions} has {\em identical scores} on a minimal pair benchmark like BLiMP \citep{warstadt-et-al-2020-blimp}, in keeping with the intuition that samples from any of these distributions are samples from the same language model, with the same passive linguistic competence.

This example illustrates a general property of minimal-pair evaluation: it abstracts away from many sources of variation that affect overall likelihood but are orthogonal to grammatical well-formedness. Perplexity, by contrast, aggregates over all factors that influence probability mass, including lexical, frequency-based, and calibration effects. In this paper, we investigate whether differences in average perplexity reflect genuine differences in grammatical sensitivity, or are dominated by other distributional factors.

\paragraph{Perplexity and emergent capabilities:} Recent studies of large LMs have identified so-called ``emergent capabilities''---tasks that LMs are poor at below a sufficient scale. \citet{DBLP:conf/nips/DuZD024} argue that this is best understood as abilities where overall model perplexity changes a small amount, but performance on that task changes a large amount. This provides additional motivation for our study---not only is it possible that the models trained on impossible languages could have exactly the same  grammatical sensitivity as measured on BLiMP tasks, but the reverse could also be the case---all models except the English model could, in principle, fail to acquire sensitivity to some key linguistic properties of their input language. 

\subsection{Generation and Iterated Learning}\label{sec:iterated}
A separate group of linguists emphasizes 
that language is shaped by cultural transmission over generations. 
On this view, sometimes called the ``Iterated Learning'' account \citep{kirby-et-al-2008-cumulative}, acquiring competence in a language is necessary, but not sufficient, for that language to persist with all its properties. The inputs to the next generation of language learners are the productions of the previous generation; accordingly, speakers must be able to reliably produce sentences in  a language for it to be transmitted to the next generation. As suggested by the temperature argument in the previous section, even perfect minimal-pair discrimination does not imply the well-formedness of generations, so in Section~\ref{sec:generation} we attempt to measure generation quality explicitly  using a pretrained LLM.

\section{Experiment 1: Minimal Pairs}
In this section, we test the grammatical sensitivity of the GPT-2 models trained by \citet{kallini_mission_2024} on ``impossible'' versions of English, by evaluating them across a suite of 67 tasks that test for different grammatical phenomena (BLiMP; \citealp{warstadt-et-al-2020-blimp}). We constructed BLiMP datasets for each impossible language by applying the corresponding perturbation rule to each sentence.\footnote{
During pre-processing, we discarded BLiMP minimal pairs that have an uneven number of tokens across the two sentences, since such pairs could become less minimally distinct in the impossible languages, due to rules which depend on linear position/number of tokens. Table \ref{tab:dataset_stats} in Appendix~\ref{app:data} lists the final dataset sizes; most datasets keep all sentences.} 
Table \ref{tab:impossible_blimp_examples} shows examples of our minimal pairs for one task.

We computed accuracy as the percentage of pairs where the log likelihood of the grammatical sentence was larger than the ungrammatical sentence. To understand how BLiMP performance degrades as the languages diverge more from human languages, we conducted two mixed effects analyses associating (1) 4-local entropy\footnote{\citet{someya-et-al-2025-information} found the strongest relationship between $m$-local entropy and perplexity for $m=4$, so we focus our analysis on this value of $m$. Calculation details for our $m$-local entropy and perplexity scores are in~Appendix~\ref{app:localperp}.} or (2) perplexity with BLiMP task accuracy, using both a fixed and random slope and intercept for each task.
 \begin{table*}[t]
    \centering
    \small
    \renewcommand{\arraystretch}{1.4}
    \begin{tabular}{l>{\raggedright}p{0.385\linewidth}p{0.388\linewidth}}
    \toprule
    \textbf{Language} & \textbf{Low Perplexity (Higher Quality)} & \textbf{High Perplexity (Lower Quality)} \\
    \midrule
    
    \textsc{English} & \textit{``Yes, that's what I'm trying to tell you.''} & \textit{``I'll do you some fancy drawing down from me.''} \\
    
    \textsc{EvenOddShuffle}\hspace{-0.25cm} & \mbox{\textit{``And what the hell are you going to do with him?''}\hspace{-0.5cm}}& \textit{``I you what wouldn said we't't I didn? ? suppose''} \\
    
    \textsc{LocalShuffle5} & \textit{``All right, let's just have a look at the camera.''} & \textit{``That's for all, is the? for what.''} \\
    \bottomrule
    \end{tabular}
    \caption{Examples of model generations between 11 and 15 tokens long in different languages, with perturbations inverted where applicable. The examples have perplexity scores below (left) and above (right) the 75th percentile of all English generations. As suggested by these examples, the top (worst) perplexity quartile for the English model still contains many grammatical (if nonsensical or unnatural) sentences. Generations for impossible languages that have perplexity at least as high are often ungrammatical. More examples for all languages are shown in Appendix~\ref{app:generation}.}
    \label{tab:generations_examples}
\end{table*}

\subsection{Overall performance}
Figure~\ref{fig:blimp} shows the performance of all models on BLiMP and the fit of our main mixed effects analysis. All models perform substantially above chance on BLiMP, with average performance over all tasks (Xs in the figure) ranging from 74.7\% accuracy for the English model to 61.5\% for \textsc{DetermShuffle}. For comparison, \citet{warstadt-et-al-2020-blimp} report that GPT-2-large obtains 80.1\% accuracy on BLiMP, compared to a human accuracy of 88.9\% (the ceiling for meaningful performance improvements). Thus, the English model, despite having 16\% as many parameters as GPT-2-large (125M vs. 775M) and being trained on 1.25\% of the data ($0.54$GB vs. $\approx40$GB), still performs fairly well, comparable with the best baseline in the BabyLM challenge (OPT-125M, 75\%; \citealp{warstadt-et-al-2023-findings}).

 Interestingly, we find that 4-local entropy captures the pattern of BLiMP scores much better than perplexity on a held-out test set ($\Delta\operatorname{AIC}=-56$). For example, \textsc{PartialReverse} has lower perplexity than \textsc{FullReverse}, but lower BLiMP accuracy and higher 4-local entropy.%
 We find a modest negative linear impact of 4-local entropy on BLiMP performance within the studied range (the solid black line in Figure~\ref{fig:blimp}: $\beta=-0.19, p\ll 0.001$, 95\% CI: $[-0.22, -0.16]$). That is, for each bit that 4-local entropy increases (somewhat more than {\em double} the difference between \textsc{English} and \textsc{LocalShuffle5}), overall BLiMP accuracy decreases by approximately 19\%. For a language like \textsc{LocalShuffle3} which is much closer to English than 1 bit, this translates into an accuracy of 68\% compared to 74\% for English.

 \subsection{Does task locality matter?}
Some BLiMP tasks may be solvable via simple local cues; such tasks could be less impacted in impossible languages than those which require more abstract generalizations. Indeed, \citeauthor{chomsky-1980-rules}'s (\citeyear{chomsky-1980-rules}) claim is precisely that rules which require structural sensitivity specifically require Universal Grammar to learn. To test whether overall BLiMP trends masked a collapse for more complex tasks, we split BLiMP tasks by their ability to be solved using local surface cues, proxied by a high accuracy score ($\geq 80\%$) by a 5-gram model in English, using the scores reported in \citet{warstadt-et-al-2020-blimp}. We refer to tasks with $\geq 80\%$ accuracy as ``local'' tasks, and others as ``non-local'' tasks. %
Table \ref{tab:dataset_stats} in Appendix~\ref{app:data} lists accuracy scores and locality for all tasks.%

Figure~\ref{fig:blimp} shows the results of including task locality as a fixed-effect in our analysis (dashed and dotted lines). The effect is significant ($p\ll0.001$), and local tasks have about 22\% higher accuracy in our analysis. However, this effect is {\em constant} regardless of 4-local entropy: adding an interaction between 4-local entropy and task locality does {\em not} result in a significantly better fit $(\chi^2(1)=0.65,p=0.42)$. That is, non-local tasks are {\em not} unusually difficult to learn in impossible languages.

At the macro-level, the results outlined in this section do not align with either extreme possibility: models neither exhibit a dramatic loss of %
grammatical sensitivity for all impossible languages, nor are they unaffected. This could indicate that grammatical sensitivity to a language might not be a suitable linking hypothesis to linguistic impossibility. In the next section, we investigate the potential of generative performance as an alternative linking hypothesis, by analysing the quality of generations sampled from \citeauthor{kallini_mission_2024}'s (\citeyear{kallini_mission_2024}) models.

\section{Experiment 2: Generative performance}\label{sec:generation}

Evaluating LM generation quality and naturalness without references is a challenging problem. One commonly used method is to use a higher quality model trained in the same language (usually on a larger dataset) to assess generations. To do this, we leverage the fact that these impossible languages can be deterministically reverted to English. After applying an inversion function to a model's generation, we evaluate it as valid English.\footnote{For inversions of the reverse languages, we removed {\em all} instances of the \sqR\space token prior to evaluation as English.} This tests the LM's ability to generate acceptable sentences, as well as how well it learned that particular perturbation rule. To evaluate a generation's acceptability, we computed the perplexity per token using a pretrained LLM (GPT2-large; \newcite{radford2019language}).

\begin{figure*}

\includegraphics[width=1\linewidth]{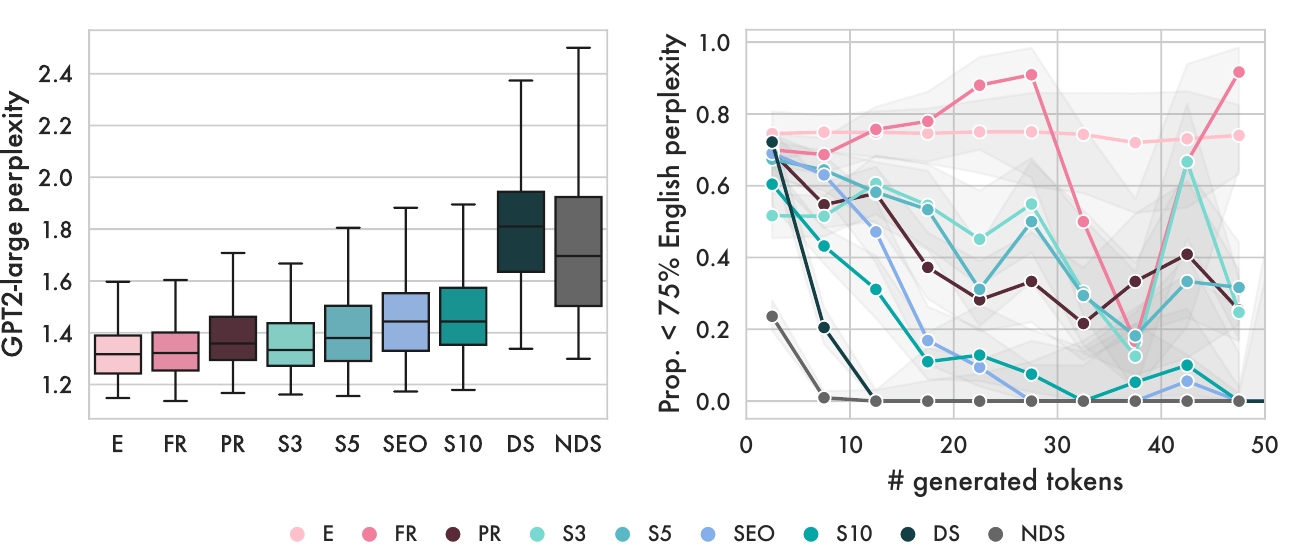}

\caption{Evaluation of generations from the impossible language models. \textbf{Left:} the perplexity under GPT2-large of generations between 11 and 20 tokens in length after inverting the ``impossible'' transformations, largely corroborating previous hierarchies of impossibility. \textbf{Right:} The proportion of generations with a perplexity below 75\% of English generations (``high-quality''), stratified by length. 
For nearly all impossible languages, substantially less than 75\% of model generations fall into this perplexity range,
with the proportion decreasing with length. \label{fig:gen}}

\end{figure*}

Although perplexity is normalized by sequence length, it still tends to have a strong correlation with the number of tokens in a sentence: early tokens tend to have higher surprisal as there is less information available to predict them and more uncertainty about the text being generated. As such, we ensure to compare perplexities across models only for matched generation lengths. We further perform an analysis based on {\em quartiles} to evaluate the proportion of generations for each model that are of a similar quality to the English model of \citet{kallini_mission_2024}. Because the generations of the English model are fairly high-quality, this serves as a simple proxy for which of the generations of the other models are of similar quality. For instance, considering a window size of 5 tokens (e.g., sentences of lengths 6-10), we compute the 25th, 50th, and 75th percentiles of English generations in terms of perplexity. Using the same window size, we then compare what proportion of generations in a particular impossible language fall {\em below} the 75th percentile/third quartile of the English model. For purposes of space and clarity, we will refer to these generations as ``high-quality.''\footnote{We chose not to use human judgments in this study, as we wanted a scalable and inexpensive evaluation for future studies to follow; still, this is an important limitation. Notably, perplexity conflates surprise due to malformedness with surprise due to sentences that convey infrequent meanings \citep{hu-et-al-2026-what}; if sentences from the impossible models use more frequent meanings
than those from the English model, this method could over-estimate their quality. However, we do not find this to be a major issue here in practice.} We generated 1000 sentences of up to 50 tokens from each model by sampling from their full trained distribution at each timestep (temperature 1). A model's beginning-of-sequence token was used as the prompt for each generation. We did not use any methods that truncate the space of considered output tokens, like top-$k$ or top-$p$ sampling, nor did we use beam search. Examples of generations with perplexities in different quartiles are shown in Table \ref{tab:generations_examples} and Appendix~\ref{app:generation}.

Figure~\ref{fig:gen} summarises the main findings of our generation experiments. 
The left-hand plot shows the distribution of perplexities for generated sentences in each language. The ranking of languages in terms of generation perplexity is similar to that based on test-set perplexity found by \citet{kallini_mission_2024} ($\rho$ = 0.93), but is even more similar to the ranking based on 4-local entropy from \citet{someya-et-al-2025-information} ($\rho$ = 0.98), diverging from test-set perplexity in the same places (chiefly \textsc{ReversePartial} and \textsc{EvenOddShuffle}).  This provides further evidence of $m$-local entropy being {\em more} predictive of model behaviour than perplexity.

The right plot in Figure~\ref{fig:gen} reveals a richer story. All models were found to generate strings of a given length at roughly the same rate (so apparent collapses are not due to a lack of sentences of that length).
All models (with the exception of \textsc{NondetermShuffle}) are able to generate a substantial (50\% or higher) percentage of high-quality sentences at a length of 5 tokens or less. However, the impossible language models tend to produce fewer high-quality generations as the number of generated tokens increase. While substantially outperforming \textsc{NondetermShuffle} for short strings, \textsc{DetermShuffle} essentially never generates high-quality strings of a length greater than 10, suggesting a failure to learn the shuffling patterns for longer sequences.

The clear outlier is \textsc{FullReverse}, which demonstrates a similar or even higher proportion of high-quality sentences compared to the base English model. The {\em higher} proportion may not indicate that the generations of \textsc{FullReverse} are actually better sentences than the English model, but rather that they express less complex/more frequent messages \citep{hu-et-al-2026-what}, highlighting the limitations of the current evaluation approach for distinguishing between models with generally high-quality generations. Nevertheless, all other impossible models have substantially fewer high-quality generations, particularly for longer sentences, suggesting substantial naturalness and acceptability issues with their generations.

\section{Discussion}
In this study, we asked what linguistic capacities LMs trained on impossible languages acquire, with the aim of identifying a plausible linking hypothesis between LM behaviour and impossibility.  Across the two experiments, we see a divergent pattern of results. Experiment 1 found that the grammatical sentitivity of an LM trained on an impossible language depends primarily on the local entropy of the language, but this effect is very modest, with LMs still achieving substantial grammatical sensitivty. In contrast, Experiment 2 showed that sampling from the learned distributions of these models yielded substantially fewer high-quality sentences than a comparable English model, especially at longer lengths. 

What implications do these divergent findings have for the impossible language debate as it connects to language models?
Our minimal pair results are amenable to multiple interpretations, largely because we lack definitive evidence about the human case. Our results do not align with generativist predictions that linguistic competency from human-scale data requires the language conform to Universal Grammar; many impossible models are almost as structurally sensitive as the base English model. However, because the English model does not achieve human level grammatical comptence, a generativist could reasonably argue that the type of grammatical learning exhibited by these models does not meaningfully bear on poverty-of-the-stimulus claims.  Nevertheless, the mounting evidence base of a steady linear climb towards grammatical competence at data levels close to the human scale (including the present results) increasingly challenges traditional poverty-of-the-stimulus views \citep{warstadt-et-al-2020-blimp, hu-et-al-2024-findings, oh-et-al-2023-transformerbased}. 

At the same time, the minimal pair results do not provide compelling evidence that the languages studied here are unlearnable by a learner with similar inductive biases to these models. The modest degradation in grammatical sensitivity shown in these results predicts that either human learners differ from current LMs in critical ways, or that humans would likewise acquire substantial grammatical sensitivity in these ``impossible'' languages---leaving open the question of why such languages do not occur. 

One possible resolution emerges from our generation results. Human language is transmitted and evolves through {\em Iterated Learning}, in which speakers learn language from the productions of a previous generation of learners \citep{kirby-et-al-2008-cumulative}. If a language is especially difficult to generate, it would be unlikely to survive transmission through generations of learners. Such a language may not be impossible to {\em learn}, but to transmit. However, this linking hypothesis faces an important challenge: {\em unlike} human language acquisition, iterated learning of language in current LMs tends to degenerate over successive generations \citep{DBLP:conf/iclr/LeBrunSO22, guo_curious_2024, shumailov-et-al-2024-aia}. 
Similar behaviour has been observed in human iterated learning experiments: even simple artificial languages degenerate over generations of human learners when there is no pressure for the language to be expressive \citep{kirby-et-al-2008-cumulative}. In humans, this pressure for expressivity can be provided by the communicative function of language: when subjects in iterated learning experiments must use their acquired language in a signalling task, the artificial language changes while maintaining its expressivity \citep{kirby-et-al-2015-compression}. While it could be possible to modify LLM training procedures to introduce an expressivity pressure \citep{smith-et-al-2024-ai}, the issue of LM degeneration nevertheless highlights critical differences between current LM training paradigms and the dynamics of human language transmission with which a successful theory must more thoroughly contend.

Additionally, our results provide new evidence for the importance of \citeauthor{someya-et-al-2025-information}'s (\citeyear{someya-et-al-2025-information}) $m$-local entropy in understanding transformer learning dynamics. While \citet{someya-et-al-2025-information} found correlations between perplexity and $m$-local entropy, we find that in both our experiments, 4-local entropy predicts LM behaviour  better than perplexity does. This contrast is striking: perplexity is a property of the {\em model being evaluated}, while 4-local entropy is {\em solely} a property of the {\em training data.} Further work is needed to understand how these findings generalize across tasks, languages, and models, and to explain why $m$-local entropy so strongly predicts generation quality and BLiMP performance.

One limitation of this study is our focus on a single transformer architecture (GPT-2 small) and set of hyperparameters. We maintain the same models as \citet{kallini_mission_2024} for the purposes of comparison, but future work should ensure findings in this area are robust to variations in training data, training duration, hyperparameters, and model specifics.

Another important limitation of our  findings is that they %
are based on a single source language: English. Prior studies extending \citet{kallini_mission_2024}'s approach to multiple languages have yielded mixed results, with ``impossible'' languages sometimes achieving {\em lower} perplexity than real human languages \citep{ziv-etal-2026-biasless, yang-et-al-2025-anything}. The interpretation of these findings is unclear, because they rely on cross-linguistic comparison of absolute perplexity, which has been shown to be unreliable across languages and datasets \citep{poelman-de-lhoneux-2026-form}. Because our approach does not depend on cross-linguistic comparisons of absolute perplexity, it may be more cross-linguistically robust. Establishing this remains a critical direction for future work, as similar confounds could still arise in practice.

\section{Conclusion}
In the emerging literature of impossible language modelling, a persistent issue has been a lack of clarity on what model behaviour needs to explain. In this paper, we have argued that a cognitively relevant theory cannot simply provide a function for identifying if a language is ``impossible,'' but must {\em also} provide a plausible linking hypothesis for why that language is unattested. Prior work had obscured the fact that models trained on impossible languages can still achieve substantial grammatical sensitivity; this fact is {\em not} aligned with {\em generative} theories of what drives language non-attestation. The picture is more promising for a transmission-oriented account, but a full account still requires substantial elaboration. Even in this simple study, though, it is clear that learning dynamics do not always align neatly with linguistic capacities of interest. We therefore urge future studies to move beyond asking how well a model fits a distribution toward asking where and why a breakdown would occur on the basis of their results.

\section*{Acknowledgements}
The authors would like to thank Iona Carslaw, Nina Gregorio, Anna Kapron-King, Oli Liu, Burin Naowarat, Yen Meng, Katarzyna Pru\'{s}, and Sydelle de Souza for their feedback during the writing process. Additionally, this work has benefitted substantially from Dagstuhl Seminar 25301 ``Linguistics and Language Models: What Can They Learn from Each Other?''---Coleman Haley would particularly like to thank the other members of the (Im)possible languages working group (Christopher Potts, Marie-Catherine de Marneffe, Katherine Demuth, Robert Frank, Juan Luis Gastaldi, Hagen Blix, Mark Johnson, Roger Levy, Kyle Mahowald, Mark Steedman, and Adina Williams) for the stimulating discussions that helped inspire this paper. 

This work was supported in part by the UKRI Centre for Doctoral Training in Natural Language Processing, funded by the UKRI (grant EP/S022481/1) and the University of Edinburgh, School of Informatics and School of Philosophy, Psychology \& Language Sciences.

\bibliography{custom, zotero}

\appendix
\begin{figure*}
    \centering
    \hspace*{0.127\linewidth}\includegraphics[width=0.86\linewidth]{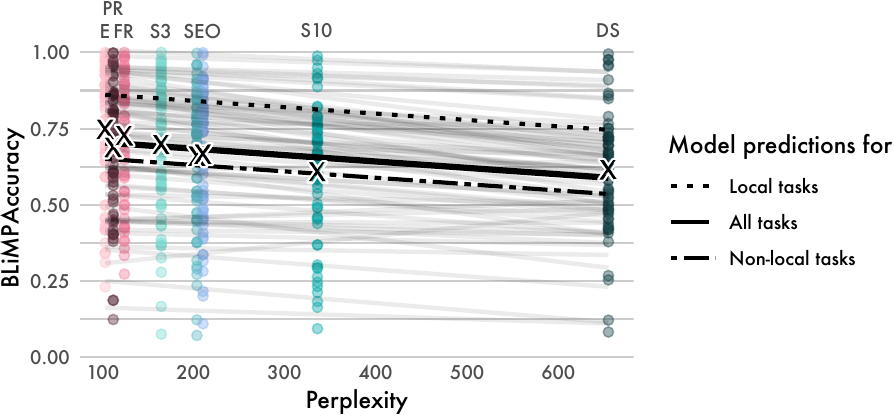}

    \caption{BLiMP task accuracy of models trained on English and impossible languages, correlated with test-set perplexity. Coloured dots and grey lines show accuracy and lines of best fit for individual tasks, with Xs indicating the mean for each language (labeled at top). The model indicates a modest, linear decline with increasing perplexity (fitted black line). This decline is insensitive to task locality: tasks that are solved well by a 5-gram model in English exhibit almost the same rate of degradation as those that are not (dotted vs. dashed line). However, we find 4-local entropy fits the data much better.}
    \label{fig:perpblimp}
\end{figure*}

\section{Details of Perplexity and 4-local Entropy}
\label{app:localperp}
\paragraph{Perplexity:} We computed the perplexity for each language on a sample of 7996 (1333 per subcorpus) sentences from perturbed versions of the BabyLM test dataset. We generated the perturbed versions by applying the perturbations to each line of the BabyLM test datasets.

\begin{table}[h]
    \centering
    \small
    \renewcommand{\arraystretch}{1.2}
    \begin{tabular}{lc}
    \toprule
    \textbf{Language} & \textbf{Perplexity} \\
    \midrule
    \textsc{English} & 102.8 \\
    \textsc{FullReverse} & 123.8 \\
    \textsc{PartialReverse} & 111.6 \\
    \textsc{LocalShuffle3} & 164.3 \\
    \textsc{LocalShuffle5} & 203.4 \\
    \textsc{EvenOddShuffle} & 210.1 \\
    \textsc{LocalShuffle10} & 335.7 \\
    \textsc{DetermShuffle} & 655.0 \\
    \textsc{NondetermShuffle} & 812.5 \\
    \bottomrule
    \end{tabular}
    \caption{Languages with their perplexity values}
    \label{tab:perplexityvals}
\end{table}

\paragraph{4-local entropy:} \citet{someya-et-al-2025-information} computed $m$-local entropy only for a subset of the languages we consider here, and on a different dataset, so we replicated their computations. We computed 4-local entropy on the 10 million sentence train set of BabyLM. This train set is a subset of the 100 million sentence set used by \citet{kallini_mission_2024}. We used \texttt{kenlm} \citet{heafield_kenlm_2011} to compute the 4-gram models over GPT-2 tokens, following \citet{someya-et-al-2025-information} in using default fallbacks for the Kneser-Ney smoothing (needed because there are no singleton unigrams after BPE tokenization).
One limitation is we did not follow \citet{kallini_mission_2024}'s sentence splitting for computing 4-local entropy, as it was very computationally expensive to run.

\begin{table}[h]
    \centering
    \small
    \renewcommand{\arraystretch}{1.2}
    \begin{tabular}{lc}
    \toprule
    \textbf{Language} & \textbf{m-local entropy} \\
    \midrule
    \textsc{English} & 2.856 \\
    \textsc{FullReverse} & 2.950 \\
    \textsc{PartialReverse} & 3.219 \\
    \textsc{LocalShuffle3} & 3.202 \\
    \textsc{LocalShuffle5} & 3.262 \\
    \textsc{EvenOddShuffle} & 3.396 \\
    \textsc{LocalShuffle10} & 3.442 \\
    \textsc{DetermShuffle} & 3.590 \\
    \textsc{NondetermShuffle} & 4.470 \\
    \bottomrule
    \end{tabular}
    \caption{Languages with their \textit{m}-local entropy value}
    \label{tab:mlocal_entropy}
\end{table}

\section{Perplexity and BLiMP Accuracy}
\label{app:perpplot}
See Figure~\ref{fig:perpblimp}.

\section{Model Generations}
\label{app:generation}
See Table~\ref{tab:expanded_generations_examples}.

\begin{table*}[h]
    \centering
    \small
    \fontsize{8}{8.5}\selectfont
    \renewcommand{\arraystretch}{1.4}
    \begin{tabular}{l p{0.4\linewidth} p{0.35\linewidth}}
    \toprule
    \textbf{Language} & \textbf{Low Perplexity (Higher Quality)} & \textbf{High Perplexity (Lower Quality)} \\
    \midrule
    
    \textsc{English} & \textit{``Well, I really don't want to be there.''} & \textit{``And you,re a poor citizen who owns it?''} \\
    \textsc{} & \textit{``I don't know what I'm gonna do this weekend, anyway.''} & \textit{``He also plays for\textbackslash xa0Derek and Wandaise.''} \\
    \textsc{} & \textit{``But I'm sorry, I didn't mean to.''} & \textit{``[Tensemble.es and applause] (Gordon)''} \\
    \hline

    \textsc{FullReverse} & \textit{``But the way I look at it, there's no doubt about it.''} & \textit{``- Take her away! - I'm sorry.!''} \\
    \textsc{} & \textit{``Oh, you're too much for me, aren't you?''} & \textit{``Watch out. - I really have to... get here.''} \\
    \textsc{} & \textit{``Yeah, because I haven't seen them for a long time.''} & \textit{``At August 1, 18051 people lived there.''} \\
    \hline

    \textsc{PartialReverse} & \textit{``I don't know, I mean, that was a great idea.''} & \textit{``Alfactory (also called the Bin Malvernis A).''} \\
    \textsc{} & \textit{``As a matter of fact, I've had nothing on it.''} & \textit{``[sister music] [Mama chuckling]''} \\
    \textsc{} & \textit{``I know what this is, but I want you to come with me.''} & \textit{``[Hoor Help!RING]ONE RING]''} \\
    \hline

    \textsc{LocalShuffle3} & \textit{``There are a lot of things that I want to do.''} & \textit{``No wonder it 'll have spoiled me now some day .''} \\
    \textsc{} & \textit{`I don't know that you were right, but I'd rather not.''} & \textit{``They were to decide that the exact sum of x ''} \\
    \textsc{} & \textit{``In 2016, she moved to Los Angeles, California, United States.''} & \textit{``You were... in mind with him,! or not?''} \\
    \hline

    \textsc{LocalShuffle5} & \textit{``You don't have to be able to get that job done, right?''} & \textit{``That's for all, is the? for what.''} \\
    \textsc{} & \textit{``No, no, I believe I'm happy to say this.''} & \textit{``Now,!'s on-by-door right here.''} \\
    \textsc{} & \textit{``All right, let's just have a look at the camera.''} & \textit{``- [! in the background] - [man...''} \\
    \hline

    \textsc{EvenOddShuffle} & \textit{``And what the hell are you going to do with him?''} & \textit{``M their and. friendsarks. theiroses come''} \\
    \textsc{} & \textit{``In 2001, he was awarded the Nobel Prize in Chemistry.''} & \textit{``Yes I,'t don we think do should. that''} \\
    \textsc{} & \textit{``You really think it's going to happen, right?''} & \textit{``I you what wouldn said we't't I didn? ? suppose''} \\
    \hline

    \textsc{LocalShuffle10} & \textit{`I've been on the bus at the time, it was my family.''} & \textit{``something'm you ask for I to thatI to.''} \\
    \textsc{} & \textit{``Now, I don't know it because I've found just about that.''} & \textit{``. place the most in the city ofThe is town''} \\
    \textsc{} & \textit{``What were you trying to do with your work to my father?''} & \textit{``- Shut up! -!!!! ( Come -.) on''} \\
    \hline

    \textsc{DetermShuffle} & \textit{-} & \textit{``the an good,, for way isIt same first?''} \\
    \textsc{} & \textit{-} & \textit{``word a know withoutDon,'t about it man?''} \\
    \textsc{} & \textit{-} & \textit{``is father like her your, then youDo?, -''} \\
    \hline

    \textsc{NondetermShuffle} & \textit{-} & \textit{`of the. a isIt in university town located city''} \\
    \textsc{} & \textit{-} & \textit{``..The people 6 2010 the city of population was''} \\
    \textsc{} & \textit{-} & \textit{``. thisBut very they a I,. when and a''} \\
    
    \bottomrule
    \end{tabular}
    \caption{Qualitative comparison of model generations between 11 and 15 tokens long in all languages used in this study. The generations, after inverting perturbations where applicable, have perplexity scores below (left) and above (right) the 75\% quantile of all English generations.}
    \label{tab:expanded_generations_examples}
\end{table*}

\section{Data Statistics}
\label{app:data}
See Table~\ref{tab:dataset_stats}.
\onecolumn

\enlargethispage{15\baselineskip}
\begingroup
\small
\setlength{\LTpre}{0pt}

\setlength{\LTpost}{0pt}
\renewcommand{\arraystretch}{1.1}
\begin{longtable}{l>{\ttfamily\fontsize{8.5}{10}\selectfont}p{205pt}rr}
    \caption{Dataset statistics and local solvability scores (Continues on next page).}\label{tab:dataset_stats} \\
\toprule

    \textbf{Phenomenon} & \textbf{\textrm{Dataset UID}} & \textbf{\# pairs} & \makecell{\textbf{Locally Solvable?} \\ \textbf{(5-gram Score)}} \\
    \midrule
    \endfirsthead
        \caption{Dataset statistics and local solvability scores (continued).} \\
\toprule

    \textbf{Phenomenon} & \textbf{\textrm{Dataset UID}} & \textbf{\# pairs} & \makecell{\textbf{Locally Solvable?} \\ \textbf{(5-gram Score)}} \\
    \midrule
\endhead
    
    ANAPHOR AGREEMENT & anaphor\_gender\_agreement & 1000 & No (44) \\
    ANAPHOR AGREEMENT & anaphor\_number\_agreement & 1000 & No (52) \\
    ARGUMENT STRUCTURE & animate\_subject\_passive & 723 & No (70) \\
    ARGUMENT STRUCTURE & animate\_subject\_trans & 555 & Yes (91) \\
    ARGUMENT STRUCTURE & causative & 581 & No (54) \\
    ARGUMENT STRUCTURE & drop\_argument & 412 & No (72) \\
    ARGUMENT STRUCTURE & inchoative & 591 & No (51) \\
    ARGUMENT STRUCTURE & intransitive & 456 & No (68) \\
    ARGUMENT STRUCTURE & passive\_1 & 644 & Yes (89) \\
    ARGUMENT STRUCTURE & passive\_2 & 607 & Yes (82) \\
    ARGUMENT STRUCTURE & transitive & 550 & No (71) \\
    BINDING & principle\_A\_c\_command & 1000 & No (58) \\
    BINDING & principle\_A\_case\_1 & 1000 & Yes (100) \\
    BINDING & principle\_A\_case\_2 & 887 & No (49) \\
    BINDING & principle\_A\_domain\_1 & 1000 & Yes (95) \\
    BINDING & principle\_A\_domain\_2 & 1000 & No (56) \\
    BINDING & principle\_A\_domain\_3 & 438 & No (52) \\
    BINDING & principle\_A\_reconstruction & 1000 & No (40) \\
    CONTROL/RAISING & existential\_there\_object\_raising & 729 & Yes (84) \\
    CONTROL/RAISING & existential\_there\_subject\_raising & 850 & No (77) \\
    CONTROL/RAISING & expletive\_it\_object\_raising & 318 & No (72) \\
    CONTROL/RAISING & tough\_vs\_raising\_1 & 933 & No (33) \\
    CONTROL/RAISING & tough\_vs\_raising\_2 & 931 & No (77) \\
    DETERMINER-NOUN AGR. & determiner\_noun\_agreement\_1 & 927 & Yes (88) \\
    DETERMINER-NOUN AGR. & determiner\_noun\_agreement\_2 & 1000 & Yes (86) \\
    DETERMINER-NOUN AGR. & determiner\_noun\_agreement\_irregular\_1 & 703 & No (53) \\
    DETERMINER-NOUN AGR. & determiner\_noun\_agreement\_irregular\_2 & 1000 & No (55) \\
    DETERMINER-NOUN AGR. & determiner\_noun\_agreement\_with\_adjective\_1 & 914 & No (52) \\
    DETERMINER-NOUN AGR. & determiner\_noun\_agreement\_with\_adj\_2 & 1000 & No (50) \\
    DETERMINER-NOUN AGR. & determiner\_noun\_agreement\_with\_adj\_irregular\_1 & 818 & No (53) \\
    DETERMINER-NOUN AGR. & determiner\_noun\_agreement\_with\_adj\_irregular\_2 & 1000 & No (55) \\
    ELLIPSIS & ellipsis\_n\_bar\_1 & 1000 & No (23) \\
    ELLIPSIS & ellipsis\_n\_bar\_2 & 593 & No (50) \\
    FILLER GAP & wh\_questions\_object\_gap & 1000 & No (53) \\
    FILLER GAP & wh\_questions\_subject\_gap & 1000 & Yes (82) \\
    FILLER GAP & wh\_questions\_subject\_gap\_long\_distance & 1000 & Yes (86) \\
    FILLER GAP & wh\_vs\_that\_no\_gap & 1000 & Yes (83) \\
    FILLER GAP & wh\_vs\_that\_no\_gap\_long\_distance & 1000 & Yes (81) \\
    FILLER GAP & wh\_vs\_that\_with\_gap & 1000 & No (18) \\
    FILLER GAP & wh\_vs\_that\_with\_gap\_long\_distance & 1000 & No (20) \\
    IRREGULAR FORMS & irregular\_past\_participle\_adjectives & 1000 & No (79) \\
    IRREGULAR FORMS & irregular\_past\_participle\_verbs & 932 & Yes (80) \\
    ISLAND EFFECTS & adjunct\_island & 1000 & No (48) \\
    ISLAND EFFECTS & complex\_NP\_island & 1000 & No (50) \\
    ISLAND EFFECTS & coordinate\_structure\_constraint\_complex\_left\_branch & 1000 & No (32) \\
    ISLAND EFFECTS & coordinate\_structure\_constraint\_object\_extraction & 1000 & No (59) \\
    ISLAND EFFECTS & left\_branch\_island\_echo\_question & 552 & Yes (96) \\
    ISLAND EFFECTS & left\_branch\_island\_simple\_question & 1000 & No (57) \\
    ISLAND EFFECTS & sentential\_subject\_island & 1000 & No (61) \\
    ISLAND EFFECTS & wh\_island & 1000 & No (56) \\
    NPI LICENSING & matrix\_question\_npi\_licensor\_present & 634 & No (1) \\
    NPI LICENSING & npi\_present\_1 & 1000 & No (47) \\
    NPI LICENSING & npi\_present\_2 & 1000 & No (47) \\
    NPI LICENSING & only\_npi\_licensor\_present & 1000 & No (57) \\
    NPI LICENSING & only\_npi\_scope & 762 & No (30) \\
    NPI LICENSING & sentential\_negation\_npi\_licensor\_present & 1000 & Yes (93) \\
    NPI LICENSING & sentential\_negation\_npi\_scope & 1000 & No (45) \\
    QUANTIFIERS & existential\_there\_quantifiers\_1 & 1000 & Yes (91) \\
    QUANTIFIERS & existential\_there\_quantifiers\_2 & 1000 & No (62) \\
    QUANTIFIERS & superlative\_quantifiers\_1 & 1000 & No (45) \\
    QUANTIFIERS & superlative\_quantifiers\_2 & 1000 & No (17) \\
    SUBJECT-VERB AGR. & distractor\_agreement\_relational\_noun & 919 & No (24) \\
    SUBJECT-VERB AGR. & distractor\_agreement\_relative\_clause & 894 & No (22) \\
    SUBJECT-VERB AGR. & irregular\_plural\_subject\_verb\_agreement\_1 & 864 & No (73) \\
    SUBJECT-VERB AGR. & irregular\_plural\_subject\_verb\_agreement\_2 & 802 & Yes (88) \\
    SUBJECT-VERB AGR. & regular\_plural\_subject\_verb\_agreement\_1 & 881 & No (76) \\
    SUBJECT-VERB AGR. & regular\_plural\_subject\_verb\_agreement\_2 & 926 & Yes (81) \\
    \bottomrule
\end{longtable}
\endgroup

\end{document}